\documentclass[10pt,twocolumn,letterpaper]{article}

\usepackage[pagenumbers]{cvpr} 

\usepackage{xspace}
\newcommand{\method}{SJD-PAC\xspace}

\usepackage{xcolor}
\usepackage{amsmath}
\usepackage{amsthm}
\usepackage{algorithm}
\usepackage{algpseudocode}
\usepackage{varwidth}
\usepackage[svgnames]{xcolor}
\usepackage{float}
\usepackage{adjustbox}
\usepackage{multirow}
\usepackage{makecell}

\usepackage{pifont}

\usepackage{accessibility}

\newcommand{\algdiff}[1]{{#1}}

\definecolor{cvprblue}{rgb}{0.21,0.49,0.74}
\usepackage[pagebackref,breaklinks,colorlinks,allcolors=cvprblue]{hyperref}

\def\paperID{30132} 
\def\confName{CVPR}
\def\confYear{2026}

\title{\method: Accelerating Speculative Jacobi Decoding via Proactive Drafting and Adaptive Continuation}

\author{%
  Jialiang Kang\textsuperscript{1} \quad
  Han Shu\textsuperscript{2} \quad
  Wenshuo Li\textsuperscript{2} \quad
  Yingjie Zhai\textsuperscript{2} \quad
  Xinghao Chen\textsuperscript{2}\thanks{Corresponding author.} \\
  \textsuperscript{1}Peking University \\
  \textsuperscript{2}Huawei Technologies \\
  \texttt{jkang@stu.pku.edu.cn} \\
  \texttt{\{han.shu,liwenshuo,zhaiyingjie,xinghao.chen\}@huawei.com}
}

\begin{document}
\maketitle
\begin{abstract}
    Speculative Jacobi Decoding (SJD) offers a draft-model-free approach to accelerate autoregressive text-to-image synthesis. However, the high-entropy nature of visual generation yields low draft-token acceptance rates in complex regions, creating a bottleneck that severely limits overall throughput. To overcome this, we introduce \method, an enhanced SJD framework. First, \method employs a proactive drafting strategy to improve local acceptance rates in these challenging high-entropy regions. Second, we introduce an adaptive continuation mechanism that sustains sequence validation after an initial rejection, bypassing the need for full resampling. Working in tandem, these optimizations significantly increase the average acceptance length per step, boosting inference speed while strictly preserving the target distribution. Experiments on standard text-to-image benchmarks demonstrate that \method achieves a $3.8\times$ speedup with lossless image quality. Code is available at https://github.com/KangJialiang/SJD-PAC.
\end{abstract}
\section{Introduction}
\label{sec:intro}

Autoregressive (AR) models for text-to-image (T2I) synthesis \cite{dalle, parti, lumina, lumina2, emu3, anole, chameleon, janus, bagel} have achieved state-of-the-art generation quality, rivaling top-tier diffusion models. These models discretize images into token sequences using a Vector Quantizer (VQ) \cite{vqgan} and frame generation as a sequence-to-sequence problem \cite{parti}. However, their primary drawback is severe inference latency, as generating a high-resolution image requires sequentially predicting thousands of tokens \cite{sjd, dalle}.

Speculative Decoding (SD) \cite{sd, lantern, lantern++} accelerates AR models by using a small, fast \emph{draft} model to propose token sequences, which a large \emph{target} model then verifies in parallel. While effective in other domains, this approach requires training and maintaining a separate draft model. More critically, standard SD techniques are largely ineffective for T2I generation \cite{lantern}. For instance, EAGLE \cite{eagle2}, despite its success in accelerating Large Language Models (LLMs), yields negligible speedups for T2I models \cite{lantern}.

This failure stems from the high-entropy nature of T2I generation. In LLMs, high sampling temperatures increase output entropy and significantly degrade the SD acceptance rate \cite{temperature_spec}. T2I models inherently exhibit this high entropy even at standard sampling temperatures \cite{t2i_entropy}. Intuitively, this means many tokens are nearly equally plausible, making it difficult for a draft model to accurately anticipate the target model's predictions.

Speculative Jacobi Decoding (SJD) \cite{sjd} offers a compelling alternative. It is both training-free and lossless, rigorously preserving the target model's output distribution. SJD re-frames generation as a fixed-point iteration, using the output distribution from the previous iteration $p^{t-1}$ as the draft for the current iteration $p^t$. Despite its elegance, SJD typically yields only a modest $\sim$2$\times$ speedup, which remains insufficient to make AR models latency-competitive.

\begin{figure}[htb]
  \centering
  \begin{subfigure}{\linewidth}
    \includegraphics{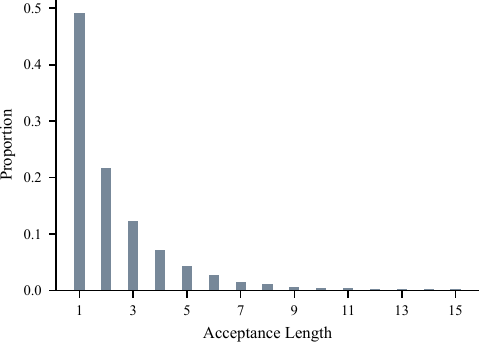}
    \caption{Frequency distribution of acceptance lengths.}
    \label{fig:histogram_dst}
    \vspace{1em}
  \end{subfigure}
  \vfill
  \begin{subfigure}{\linewidth}
    \includegraphics{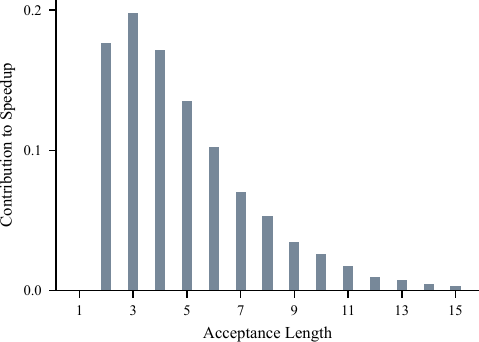}
    \caption{Proportional contribution to total speedup by acceptance length.}
    \label{fig:histogram_contri}
  \end{subfigure}
  \caption{Analysis of acceptance lengths in SJD \cite{sjd}. The speedup shown in (b) is derived entirely from additionally accepted tokens (acceptance length $> 1$).}
  \label{fig:histogram}
\end{figure}

To understand this limitation, we analyze SJD's practical performance by examining the distribution of accepted tokens per forward step. As shown in \cref{fig:histogram_dst}, this distribution is highly skewed and long-tailed. In nearly 50\% of forward passes, SJD accepts only a single token. This occurs when the first draft token fails verification, triggering a resample and yielding zero acceleration for that step. Consequently, as \cref{fig:histogram_contri} illustrates, these single-token steps contribute nothing to the overall speedup (which is derived solely from extra accepted tokens, i.e., acceptance length $> 1$). The average $\sim$2$\times$ speedup is thus disproportionately driven by a small fraction of steps that successfully accept numerous tokens in a single pass.

This observation naturally raises two critical questions:
\begin{enumerate}
  \item \textbf{Mitigating Low-Efficiency Steps:} How can we reduce the frequency of single-token acceptances to minimize their drag on overall speedup?
  \item \textbf{Maximizing Acceptance Length:} How can we increase the number of successfully verified tokens in a single pass to amplify their contribution to acceleration?
\end{enumerate}
\vspace{.5em}

In this paper, we propose \method, an enhanced SJD framework that directly addresses both challenges. Crucially, \method is entirely training-free and rigorously lossless.

To \textit{mitigate low-efficiency steps}, we introduce \textbf{Proactive Drafting (PD)}. This strategy targets the root cause of single-token acceptances: the context mismatch caused by a rejection. When a token at position $i$ is rejected, the context for all subsequent proposals $x_{j > i}$ becomes invalid, often triggering cascading rejections in the subsequent iteration. PD addresses this by modifying the drafting phase. Instead of proposing a single sequence, PD proactively constructs multiple diverse proposal paths rooted at the newly resampled position. By providing several alternatives, PD increases the likelihood that at least one path remains valid during verification, directly suppressing the spike in single-token acceptances.

To \textit{maximize acceptance length}, we introduce \textbf{Adaptive Continuation (AC)}. This technique modifies the standard SJD verification loop, which typically terminates upon the first rejection. AC, in contrast, eliminates this hard termination. When a rejection occurs at position $i$, AC resamples only the failed token $x_i^t$ and continues the verification process for all subsequent positions $j > i$. This mechanism allows AC to preserve valid tokens beyond the initial point of failure, discarding only the rejected ones. This creates a more stable initial state for the next Jacobi iteration while maintaining rigorous losslessness.

Our main contributions are as follows:
\begin{itemize}
  \item We identify and analyze the skewed acceptance-length distribution of SJD in T2I models, pinpointing it as the primary performance bottleneck.
  \item We introduce \method, a novel, training-free, and lossless SJD variant incorporating two new techniques: Proactive Drafting (PD) to mitigate cascading rejections, and Adaptive Continuation (AC) to preserve valid tokens beyond the initial failure.
  \item We empirically demonstrate that \method achieves state-of-the-art acceleration for lossless T2I generation, significantly outperforming the original SJD \cite{sjd} and recent lossy alternatives \cite{gsd, sjd2}.
\end{itemize}
\section{Related Works}
\label{sec:related}

\subsection{Autoregressive Text-to-Image Generation}
Early autoregressive T2I models \cite{dalle} utilized a two-stage pipeline. First, an image tokenizer, such as a VQ-VAE \cite{vqvae}, discretized the image into a 1D sequence of tokens. A large decoder-only Transformer \cite{attention} then generated this sequence autoregressively, conditioned on a text prompt. Parti \cite{parti} exemplified this paradigm, scaling the Transformer to billions of parameters to achieve high-fidelity synthesis. More recently, the field has shifted toward unified, end-to-end architectures \cite{lumina, lumina2, emu3, anole, chameleon, janus, bagel}. By jointly co-training the tokenizer and Transformer on vast mixed-modal datasets within a single framework, these models build a richer contextual understanding and achieve state-of-the-art generation quality. Despite these architectural advancements, the fundamental autoregressive bottleneck remains: tokens must be generated sequentially during inference, resulting in significant latency.

\subsection{Speculative Decoding}
Speculative Decoding (SD) \cite{sd} has emerged as an effective acceleration strategy for Large Language Models (LLMs). It typically employs a small draft model to generate candidate tokens, which are subsequently verified in parallel by a large target model. This core concept has been widely extended: SpecInfer \cite{specinfer} utilizes tree-based drafting for more efficient verification, while ViSpec \cite{vispec} adapts the principle for Vision-Language Models (VLMs).

In the context of T2I models, LANTERN \cite{lantern} addresses the inherently low acceptance rates of image tokens by proposing a relaxed SD framework that accepts plausible, albeit non-identical, proposals. LANTERN++ \cite{lantern++} further enhances this approach using static tree drafting. Concurrently, a distinct, training-free paradigm was introduced with Speculative Jacobi Decoding (SJD) \cite{sjd}. SJD leverages a fixed-point iteration mechanism, enabling the target model to generate its own drafts. This achieves lossless acceleration without requiring an auxiliary draft model. Subsequent methods, such as Grouped Speculative Decoding (GSD) \cite{gsd} and SJD2 \cite{sjd2}, have explored variations of this approach, though often at the cost of reintroducing training requirements or compromising the lossless guarantee.

\begin{figure}[htb]
    \centering
    \includegraphics{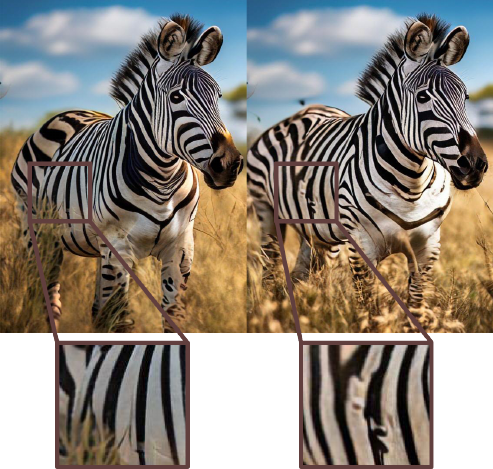}
    \caption{High sensitivity of AR T2I models to token perturbations. \textbf{Left}: Original generation. \textbf{Right}: Replacing a single token (0.04\% of total tokens) during the autoregressive process causes a severe visual artifact (red box).}
    \label{fig:token_artifact}
\end{figure}

While some relaxed SD approaches operate on the premise that substituting specific tokens with similar alternatives does not necessarily degrade perceptual quality \cite{gsd}, image generation can be exceedingly sensitive to token-level perturbations. As illustrated in \cref{fig:token_artifact}, modifying a single token during generation can introduce severe visual artifacts into the final output. Consequently, maintaining a lossless guarantee remains critical for the reliable application of SD in T2I models.

\section{Preliminaries}
\label{sec:prelim}

\subsection{Notations}
\label{sec:notations}
In this paper, we denote the target T2I model to be accelerated as $M$. Let $p(\cdot\mid\cdot)$ be its output probability distribution. A proposal distribution $q(\cdot\mid\cdot)$ is used to generate draft tokens. If a separate draft model is utilized, $q$ represents its output distribution; in training-free methods like SJD \cite{sjd}, $q$ is the distribution derived from cached, stale logits from a previous iteration of $M$. We use lowercase for individual tokens, e.g., $x$, and uppercase for sequences, e.g., $X$. Subscripts denote token positions and superscripts denote iterations; thus, $x_i^t$ is the token at position $i$ in iteration $t$. We use $X_{i:j}$ to represent the subsequence $(x_i, \dots, x_j)$ and $X_{<i}$ for the prefix $(x_1, \dots, x_{i-1})$.

\subsection{Rejection Sampling}
\label{sec:rejection_sampling}
Rejection sampling~\cite{rejection_sampling} is a Monte Carlo method for drawing samples from a target distribution $p$ using a proposal distribution $q$. The technique guarantees that accepted samples are distributed exactly according to $p$, provided that $p(x) \le c \cdot q(x)$ holds for all $x$ and a known constant $c \ge 1$. Crucially, $p$ and $q$ are not required to be conditional on the same context or prefix~\cite{rejection_sampling, sjd}. In speculative decoding, a sample $x \sim q$ is drawn and accepted if $u \le p(x) / q(x)$, where $u \sim U(0, 1)$. If $x$ is rejected, a new token is sampled from the normalized residual distribution $p'_{\text{res}}(x) \propto \max(p(x) - q(x), 0)$. For brevity, we will henceforth use $\max(p(x) - q(x), 0)$ to denote this calibrated distribution, omitting the normalization factor.

\subsection{Speculative Jacobi Decoding}
SJD~\cite{sjd} treats AR generation as a fixed-point iteration problem. It operates on a token sequence $X = (x_1, \dots, x_L)$ within a fixed-length Jacobi window of size $L$. At the start of iteration $t$, the inputs are the token sequence $X^{t-1}$ and its corresponding stale output distributions $P^{t-1} = M(X^{t-2})$ cached from the previous iteration. A single forward pass on $X^{t-1}$ yields the new target distributions $P^t = M(X^{t-1})$.

The core SJD verification loop then proceeds sequentially from $i=1$ to $L$. At each position $i$:
\begin{enumerate}
  \item \textbf{Target distribution:} $p_i = P^t(\cdot \mid X_{<i}^{t-1})$.
  \item \textbf{Proposal distribution:} $q_i = P^{t-1}(\cdot \mid X_{<i}^{t-2})$.
  \item \textbf{Draft token:} $x_i^{t-1}$, generated in the previous iteration.
  \item \textbf{Verification:} $x_i^{t-1}$ is accepted or rejected via rejection sampling using $p_i$ and $q_i$.
\end{enumerate}
\vspace{.5em}

If $x_i^{t-1}$ is accepted, we set $x_i^t \leftarrow x_i^{t-1}$ and proceed to position $i+1$. If $x_i^{t-1}$ is rejected, a new token $x_i^t$ is sampled from the calibrated distribution $\max(p_i-q_i, 0)$, and the verification loop terminates. Following termination at position $i$, the remaining tokens $X_{i+1:L}^t$ are sampled from their respective target distributions $p_j(\cdot \mid X_{<j}^{t-1})$ for $j = i+1, \dots, L$. Finally, $i$ new tokens are initialized and appended to maintain the window size $L$, and the process advances to iteration $t+1$.
\section{Method}
\label{sec:method}
\begin{figure}[htb]
    \centering
    \includegraphics{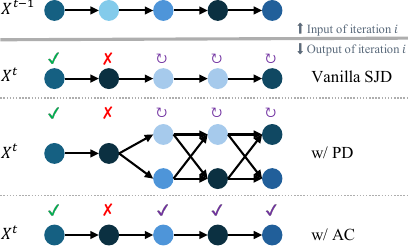}
    \caption{A conceptual overview of our proposed components compared to Vanilla SJD, illustrating how each method handles a token rejection (\textcolor{red}{red \ding{55}}) at iteration $t$.
        \textbf{Vanilla SJD} \textbf{(Top)}: A rejection at position 2 forces a \texttt{break}, and the entire subsequent sequence is resampled (\textcolor{violet}{violet $\mathbf{\circlearrowright}$}).
        \textbf{w/ PD} \textbf{(Middle)}: After a rejection, Proactive Drafting builds a diverse $K$-ary tree at the rejection point to increase the probability of acceptance in the subsequent iteration.
        \textbf{w/ AC} \textbf{(Bottom)}: After a rejection, Adaptive Continuation \textit{continues} the verification loop, preserving subsequent valid tokens (\textcolor{violet}{violet \ding{51}}) instead of terminating.}
    \label{fig:component_overview}
\end{figure}

While SJD \cite{sjd} is efficient and lossless, our analysis in \cref{sec:intro} reveals that its performance is bottlenecked by frequent rejections and short acceptance lengths. In this section, we introduce our proposed method, \method, which incorporates two key innovations: Proactive Drafting (PD) and Adaptive Continuation (AC). These techniques work synergistically to reduce the frequency of rejections and increase the number of tokens accepted per forward pass, thereby enhancing overall generation speed without sacrificing output quality. A conceptual illustration of how PD and AC modify the standard SJD rejection process is provided in \cref{fig:component_overview}.

\subsection{Proactive Drafting (PD)}
\label{sec:pd}
To \textbf{mitigate low-efficiency steps}, specifically the frequent single-token acceptances, we first modify the \textit{proposal generation} process that follows a rejection. In standard SJD \cite{sjd}, a rejection at position $i$ during iteration $t$ creates a context mismatch. After the new token $x_i^t$ is resampled, the standard SJD procedure generates the remainder of the sequence $X_{i+1:L}^t$ by sampling from the target distributions $p_j(\cdot \mid X_{<j}^{t-1})$, where $j = i+1, \dots, L$. The problem is that these distributions are conditioned on a stale prefix $X_{<j}^{t-1}$, which contains the rejected token $x_i^{t-1}$ rather than the corrected $x_i^t$. Consequently, this new proposal $X_{i+1:L}^t$ is misaligned with its own context. When used as the draft in the next iteration $t+1$, the tokens $X_{i+1:L}^t$ are highly likely to be rejected immediately at $i+1$, leading to a cascading spike of single-token acceptances.

PD addresses this by proactively constructing a new, more diverse proposal $X^t$ the moment a rejection occurs at $i$. After the new token $x_i^t$ is resampled, we build this new proposal rooted at $x_i^t$:
\begin{enumerate}
    \item \textbf{Shallow, Wide Tree:} For positions $j$ from $i+1$ to $i+D$ (e.g., $D=3$), we sample $K$ candidate tokens (e.g., $K=4$) without replacement from the target distribution $p(\cdot \mid X_{<j}^{t-1})$. This forms a $K$-ary tree of depth $D$, providing a diverse set of local candidates.
    \item \textbf{Chain Extension:} For all subsequent positions $j > i+D$, we extend one of these paths (e.g., the first path, $k=1$) as a single proposal chain up to the full length $L$, by sampling autoregressively from $p(\cdot \mid X_{<j}^{t-1})$.
\end{enumerate}
\vspace{.5em}

This new hybrid proposal, a shallow tree followed by a long chain, is then fed into iteration $t+1$. By providing $K$ diverse options at the critical, post-rejection local boundary, PD increases the likelihood that at least one path will remain valid, directly mitigating the single-token acceptance spike. Unlike tree-based speculative decoding \cite{specinfer}, our tree is built only locally and is conditioned on the potentially stale context $X^{t-1}$. This avoids the need for multiple model forward passes.

\subsection{Adaptive Continuation (AC)}
\label{sec:ac}
To \textbf{maximize the acceptance length}, we modify the verification loop of standard SJD. In SJD, the verification at iteration $t$ terminates upon the first rejection at position $i$, discarding all subsequent proposal tokens $X_{i+1:L}^{t-1}$. This is inefficient, as many tokens might remain valid under the shifted context. Adaptive Continuation (AC) eliminates this hard \texttt{break} statement. Upon a rejection at $i$, AC resamples $x_i^t \sim \max(p_i - q_i, 0)$ and continues verifying the remaining positions $j = i+1, \dots, L$.

\begin{figure}[htb]
    \centering
    \includegraphics{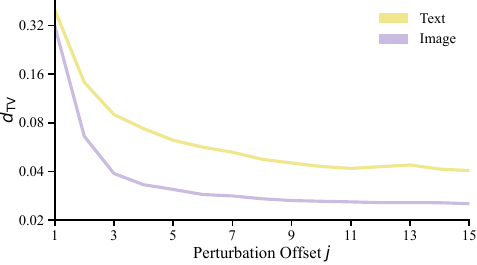}
    \caption{Total Variation distance $d_{\text{TV}}$ for text and image generation with respect to the perturbation offset $j$.}
    \label{fig:tv_perturbation}
\end{figure}

For $j > i$, the updated prefix $X_{<j}^t$ renders the pre-computed draft and target distributions $(p_j, q_j)$ stale. Recomputing them would negate the speedup. However, utilizing these stale distributions remains highly effective due to the strong \textit{locality} of image tokens. To empirically validate this, we use MS-COCO~\cite{coco} and prompt Lumina-mGPT~\cite{lumina} to generate both images and text stories. By individually masking the KV-cache at a preceding position with an offset $j$ (i.e., at position $i-j$), we measure the Total Variation (TV) distance of the output distribution at position $i$ before and after the perturbation. As shown in \cref{fig:tv_perturbation}, the TV distance $d_{\text{TV}}$ for image tokens rapidly approaches zero as the distance from the perturbed context increases. This contrasts sharply with text generation, which remains highly sensitive to distant context changes.

Since the acceptance probability of a single draft token is $1 - d_{\text{TV}}(p, q)$~\cite{spectr}, standard rejection sampling naturally preserves tokens when $d_{\text{TV}}$ is low. Specifically, the probability that a token at position $i$ remains unchanged is $1 - d_{\text{TV}}(p^{t-1}_i, p^{t}_i)$. In contrast, if SJD were to independently resample the same token, the probability would be $\sum_{\mathcal{V}} p^{t-1}_i p^{t}_i \leq \exp\left(-\frac{H_2(p^{t-1}) + H_2(p^{t})}{2}\right)$, where $H_2$ denotes the order-2 R{\'e}nyi entropy~\cite{fid}. Given the high entropy of image distributions, this resampling probability is negligible ($<0.01$). By evaluating the remaining tokens using the stale distributions $(p_j, q_j)$ via standard rejection sampling, AC effectively exploits the asymptotic behavior observed in \cref{fig:tv_perturbation} to maximize token retention, achieving a retention probability $>0.7$:
\begin{itemize}
    \item \textbf{Acceptance:} Set $x_j^t \leftarrow x_j^{t-1}$, preserving the token.
    \item \textbf{Rejection:} Sample a new token $x_j^t \sim \max(p_j - q_j, 0)$.
\end{itemize}

This process yields a full-length sequence $X^t$ with minimal deviation from $X^{t-1}$, providing a stable initialization for iteration $t+1$. Crucially, AC remains rigorously lossless. The tail sequence $X_{>i}^t$, although generated using stale distributions, serves only as the draft for the next iteration, where it will be verified against updated, exact target distributions.

\subsection{Full Algorithm}
\begin{algorithm}[htb]
    \caption{\method (Ours)}
    \label{alg:sjd-pac}
    \begin{algorithmic}[1]
        \State $X^{t-1}$: draft tokens, $P^{t-1}$: draft probs
        \State $P^t \gets \Call{ParallelForward}{X^{t-1}}$
        \State $X^t \gets [\,]$
        \State \algdiff{$\mathtt{first\_rej\_idx} \gets -1$}
        \For{\algdiff{$i = 1 \to L$}} \Comment{\algdiff{Adaptive Continuation Loop}}
        \State $p_i \gets P^t_i$; $q_i \gets P^{t-1}_i$; $x_i^{t-1} \gets X^{t-1}_i$
        \State $r \sim U[0, 1]$
        \If{$r < \min(1, \frac{p_i}{q_i})$}
        \State $x_i^t \gets x_i^{t-1}$ \Comment{Accept}
        \Else
        \State $x_i^t \gets \Call{Sample}{\max(p_i - q_i, 0)}$ \Comment{Reject}
        \If{\algdiff{$\mathtt{first\_rej\_idx} = -1$}}
        \State \algdiff{$\mathtt{first\_rej\_idx} \gets i$}
        \EndIf
        \EndIf
        \EndFor \Comment{\algdiff{No break}}

        \If{\algdiff{$\mathtt{first\_rej\_idx} \neq -1$}} \Comment{\algdiff{Proactive Drafting}}
        \State \algdiff{$X^t \gets \Call{PD}{X^t, \mathtt{first\_rej\_idx}}$}
        \EndIf
        \State \Return $X^t, P^t$
    \end{algorithmic}
\end{algorithm}

Our full algorithm, \method, is detailed in \cref{alg:sjd-pac}. It integrates the two aforementioned components: Adaptive Continuation (AC) and Proactive Drafting (PD). The algorithm begins by performing a parallel forward pass to obtain the target probabilities $P^t$ (Line 2). We initialize a variable $\mathtt{first\_rej\_idx}$ to track the position of the first mismatch (Line 4).

In the AC Loop (Lines 5--17), unlike standard SJD which terminates on the first rejection, our loop always continues to verify all $L$ tokens. During this pass, standard acceptance (Line 10) or rejection sampling (Line 12) is performed for each token. If a token is rejected, we check if it is the first rejection; if so, we record its position in $\mathtt{first\_rej\_idx}$ (Lines 13-15) and continue the loop without breaking.

After the loop, we check if any rejections occurred (Line 19). If $\mathtt{first\_rej\_idx}$ is not -1, we activate the Proactive Drafting (PD) mechanism (Line 20). The PD function leverages the target distribution $p$ to generate diversified token sequences starting from the position of the first rejection.

\method is rigorously lossless, as both the AC and PD components preserve the underlying target distribution. The framework is also training-free and model-agnostic.

\section{Experiments}
\label{sec:exp}

\subsection{Experimental Setup}

\paragraph{Models and Benchmarks.}
We evaluate our method on two text-to-image models: Lumina-mGPT~\cite{lumina} and Emu3~\cite{emu3}.
For Lumina-mGPT, we use the standard 7B model at a $768 \times 768$ resolution.
Emu3 is an 8B model that generates $720 \times 720$ images.
Following standard practice~\cite{lumina}, we use a top-2000 logit sampler, a temperature of 1, and a classifier-free guidance (CFG) weight of 3 for inference.
We conduct evaluations on two standard benchmarks: the MS-COCO 2017 validation set~\cite{coco} and the PartiPrompts dataset~\cite{parti_prompts}.

\paragraph{Baselines and Implementation.}
We compare \method against several strong decoding baselines. These include lossless methods, namely EAGLE-2~\cite{eagle2} and Speculative Jacobi Decoding (SJD)~\cite{sjd}, as well as high-performing lossy methods, including LANTERN++~\cite{lantern++}, GSD~\cite{gsd}, and SJD2~\cite{sjd2}. Our implementation of EAGLE-2 is adapted from the LANTERN codebase~\cite{lantern}. For all other baselines, we use the official parameters and checkpoints from their respective papers. To ensure a fair comparison, if our reproduced metric for a baseline falls below its reported score, we report the published value. For our method, we use a draft tree with depth $D=3$ and $K=4$ branches per node, and set the verification window size to $L=64$.

\paragraph{Evaluation Metrics.}
We evaluate both generation quality and decoding efficiency.
\textbf{Quality:} We use the Fr\'echet Inception Distance (FID)~\cite{fid, gans} and CLIP-Score~\cite{openclip,clipscore} to assess visual fidelity. Both metrics are reported on the MS-COCO validation set. Since the PartiPrompts dataset lacks ground-truth images, we report only the CLIP-Score for this benchmark.
\textbf{Speed:} We report two metrics.
\begin{itemize}
    \item \textbf{Step Compression Ratio}, defined as:
          $$\mathcal{S} = \frac{\text{generated tokens}}{\text{decoding steps}},$$
          which measures theoretical acceleration.
    \item \textbf{End-to-end Speedup}, which measures the practical wall-clock speedup against standard autoregressive decoding.
\end{itemize}

\subsection{Main Results}
\label{sec:main_results}
\Cref{tab:main_results} presents the main speed and quality comparisons on the MS-COCO 2017~\cite{coco} and PartiPrompts~\cite{parti_prompts} datasets using Lumina-mGPT~\cite{lumina} and Emu3~\cite{emu3}. As shown in the table, \method achieves state-of-the-art acceleration across all configurations while rigorously maintaining baseline image quality.

For Lumina-mGPT, \method achieves the highest acceleration, recording a $\mathbf{4.51\times}$ step compression and a $\mathbf{3.80\times}$ wall-clock speedup on MS-COCO. This significantly surpasses other lossless methods like EAGLE~\cite{eagle2} ($2.10\times$) and SJD~\cite{sjd} ($2.05\times$). Crucially, this speedup incurs zero quality degradation: our FID and CLIP-Score are virtually identical to the original model (30.69 vs. 30.79 and 31.21 vs. 31.31, respectively). Notably, our lossless acceleration even exceeds that of lossy counterparts. For instance, while GSD~\cite{gsd} achieves a $3.62\times$ latency speedup, it suffers a notable quality drop (33.12 FID). In contrast, \method is demonstrably faster ($3.80\times$) while preserving high fidelity.

The results on Emu3~\cite{emu3} reinforce this trend. \method again achieves the highest speedup among all lossless methods ($\mathbf{3.25\times}$ on MS-COCO) while preserving the baseline generation quality (FID: 31.10 vs. 31.12). Although SJD2 reports a higher step compression ratio ($5.62\times$), it requires double our window length (128 vs. 64). This introduces significant computational overhead that limits its practical acceleration, resulting in a much lower wall-clock speedup ($2.54\times$).

We observe a consistent pattern on the PartiPrompts dataset. For both Lumina-mGPT and Emu3, \method delivers the best lossless wall-clock speedup ($\mathbf{3.97\times}$ and $\mathbf{3.51\times}$, respectively) and the highest step compression in the lossless category. Across all cases, our method preserves the baseline CLIP-Score, outperforming methods like LANTERN++~\cite{lantern++}, GSD~\cite{gsd}, and SJD2~\cite{sjd2}, which all exhibit quality degradation. While the lossy GSD~\cite{gsd} posts a high latency speedup ($4.65\times$) on Lumina-mGPT, \method ($3.97\times$) remains highly competitive without the associated quality drop (our 32.07 vs. GSD's 31.25).

These quantitative results demonstrate that \method establishes a new state-of-the-art. It outperforms all existing lossless techniques in both step reduction and latency speedup, yielding acceleration competitive with lossy methods, all without compromising the foundation model's generation quality.

\begin{table*}[htb]
    \centering
    \caption{
        Main results comparing our \method method against baselines on acceleration and image quality.
        We report Step Compression (\textuparrow), wall-clock Latency speedup (\textuparrow), FID (\textdownarrow), and CLIP-Score (\textuparrow) on the Lumina-mGPT and Emu3 models.
        Evaluations are conducted on the MS-COCO 2017 and PartiPrompts benchmarks.
        \textbf{Bold} indicates the highest speedups in each category.
    }
    \label{tab:main_results}
    \renewcommand{\arraystretch}{1.05}
    \setlength{\tabcolsep}{.98em}
    \begin{tabular}{l c c c c c c}
        \toprule
        \multirow{2}{*}{\textbf{Configuration}}                  & \multirow{2}{*}{\textbf{\makecell{Training-                                                                                                          \\Free}}} & \multirow{2}{*}{\textbf{Lossless}} & \multicolumn{2}{c}{\textbf{Acceleration} (↑)} & \multicolumn{2}{c}{\textbf{Image Quality}}
        \\
        \cmidrule(lr){4-5} \cmidrule(lr){6-7}
                                                                 &                                             &           & \textbf{Step}         & \textbf{Latency}      & \textbf{FID} (↓) & \textbf{CLIP-Score} (↑) \\
        \midrule
        \multicolumn{7}{l}{\textbf{MS-COCO 2017}}                                                                                                                                                                       \\
        Lumina-mGPT \cite{lumina}                                & \ding{51}                                   & \ding{51} & $1.00\times$          & $1.00\times$          & 30.79            & 31.31                   \\
        w/ EAGLE \cite{eagle2}                                   & \ding{55}                                   & \ding{51} & $2.94\times$          & $2.10\times$          & 30.68            & 31.73                   \\
        w/ SJD \cite{sjd}                                        & \ding{51}                                   & \ding{51} & $2.22\times$          & $2.05\times$          & 31.13            & 31.33                   \\
        w/ $\text{LANTERN++}_{k=10, \lambda=2}$ \cite{lantern++} & \ding{55}                                   & \ding{55} & $3.19\times$          & $2.28\times$          & 29.96            & 30.11                   \\
        w/ $\text{GSD}_{G=10}$ \cite{gsd}                        & \ding{51}                                   & \ding{55} & $3.39\times$          & $3.62\times$          & 33.12            & 31.25                   \\
        w/ SJD2 \cite{sjd2}                                      & \ding{55}                                   & \ding{55} & $4.02\times$          & $2.81\times$          & 31.40            & 31.80                   \\
        \textbf{w/ \method (Ours)}                               & \ding{51}                                   & \ding{51} & $\mathbf{4.51\times}$ & $\mathbf{3.80\times}$ & 30.69            & 31.21                   \\

        \midrule

        Emu3 \cite{emu3}                                         & \ding{51}                                   & \ding{51} & $1.00\times$          & $1.00\times$          & 31.12            & 31.05                   \\
        w/ SJD \cite{sjd}                                        & \ding{51}                                   & \ding{51} & $2.32\times$          & $2.01\times$          & 30.74            & 30.95                   \\
        w/ SJD2 \cite{sjd2}                                      & \ding{55}                                   & \ding{55} & $\mathbf{5.62\times}$ & $2.54\times$          & 31.50            & 30.40                   \\
        \textbf{w/ \method (Ours)}                               & \ding{51}                                   & \ding{51} & $4.31\times$          & $\mathbf{3.25\times}$ & 31.10            & 30.99                   \\

        \midrule
        \midrule

        \multicolumn{7}{l}{\textbf{PartiPrompts}}                                                                                                                                                                       \\
        Lumina-mGPT \cite{lumina}                                & \ding{51}                                   & \ding{51} & $1.00\times$          & $1.00\times$          & -                & 32.01                   \\
        w/ EAGLE \cite{eagle2}                                   & \ding{55}                                   & \ding{51} & $2.86\times$          & $2.01\times$          & -                & 32.01                   \\
        w/ SJD \cite{sjd}                                        & \ding{51}                                   & \ding{51} & $2.28\times$          & $2.13\times$          & -                & 32.06                   \\
        w/ $\text{LANTERN++}_{k=10, \lambda=2}$ \cite{lantern++} & \ding{55}                                   & \ding{55} & $3.02\times$          & $2.10\times$          & -                & 31.07                   \\
        w/ $\text{GSD}_{G=50}$ \cite{gsd}                        & \ding{51}                                   & \ding{55} & $3.76\times$          & $\mathbf{4.65\times}$ & -                & 31.25                   \\
        w/ SJD2 \cite{sjd2}                                      & \ding{55}                                   & \ding{55} & $3.82\times$          & $2.51\times$          & -                & 31.54                   \\
        \textbf{w/ \method (Ours)}                               & \ding{51}                                   & \ding{51} & $\mathbf{4.62\times}$ & $3.97\times$          & -                & 32.07                   \\

        \midrule

        Emu3 \cite{emu3}                                         & \ding{51}                                   & \ding{51} & $1.00\times$          & $1.00\times$          & -                & 31.85                   \\
        w/ SJD \cite{sjd}                                        & \ding{51}                                   & \ding{51} & $2.35\times$          & $2.11\times$          & -                & 31.65                   \\
        w/ SJD2 \cite{sjd2}                                      & \ding{55}                                   & \ding{55} & $\mathbf{4.72\times}$ & $2.04\times$          & -                & 31.23                   \\
        \textbf{ w/ \method (Ours)}                              & \ding{51}                                   & \ding{51} & $4.59\times$          & $\mathbf{3.51\times}$ & -                & 31.87                   \\
        \bottomrule
    \end{tabular}
\end{table*}

\subsection{Ablation Study}
This section presents an ablation study evaluating our proposed components, Proactive Drafting (PD) and Adaptive Continuation (AC), along with the key hyperparameters of the PD mechanism. All experiments use the Lumina-mGPT model on the MS-COCO dataset.

\subsubsection{Effectiveness of Proposed Components}
We incrementally add our components to the SJD baseline to quantify their impact, as shown in \cref{tab:ablation_components}. The baseline SJD with a window length $L=32$ achieves $2.31\times$ step compression. Integrating Proactive Drafting (PD) increases this to $2.71\times$, highlighting the benefit of drafting a $K$-ary tree over a simple chain, which allows for more potential acceptance paths. Adding Adaptive Continuation (AC), which stabilizes the sequence after rejections (see \cref{sec:ac}), further improves compression to $3.52\times$.

With AC equipped, the $L=32$ window becomes a bottleneck as tokens stabilize faster. Expanding the window length to $L=64$ fully leverages our method, achieving our main result of \textbf{$4.51\times$} step compression. While larger window sizes ($L>64$) yield marginal gains, they incur significant computational overhead that negates practical speedups on our hardware. Therefore, we select $L=64$ as the optimal default to balance efficiency and speed. These results confirm that both PD and AC are crucial for performance.

\begin{table}[htb]
    \centering
    \caption{Ablation study on the contribution of each component (PD, AC) and the effect of window length ($L$).}
    \label{tab:ablation_components}
    \setlength{\tabcolsep}{.74em}
    \begin{tabular}{l c c}
        \toprule
        \textbf{Method} & \textbf{Window Len} & \textbf{Step Compression} (↑) \\
        \midrule
        Baseline        & 32                  & $2.31\times$                  \\
        + PD            & 32                  & $2.71\times$                  \\
        + PD + AC       & 32                  & $3.52\times$                  \\
        + PD + AC       & \textbf{64}         & $\mathbf{4.51\times}$         \\
        \bottomrule
    \end{tabular}
\end{table}

\subsubsection{Proactive Drafting Hyperparameters}
We investigate the two key hyperparameters of PD: the $K$-ary factor $K$ and the tree depth $D$. As shown in \cref{tab:ablation_hyperparams}, when fixing $D=3$, increasing $K$ from 2 to 4 yields consistent improvements ($4.23\times$ to $4.51\times$). However, performance slightly drops at $K=5$ ($4.47\times$). Similarly, when fixing $K=4$, $D=3$ ($4.51\times$) outperforms both a shallower $D=2$ ($4.34\times$) and a deeper $D=4$ ($4.49\times$).

This trade-off occurs because the total number of verified tokens per step, i.e., the computational budget, is fixed in this setup. Allocating more tokens to the drafting tree by increasing $K$ or $D$ inherently shortens the subsequent drafting chain. Maintaining this fixed budget prevents the computational overhead from becoming compute-bound—an issue that diminishes wall-clock acceleration in methods like SJD2~\cite{sjd2} due to excessively long windows (see \cref{sec:main_results}). Consequently, we identify \textbf{$K=4, D=3$} as the optimal configuration.

\begin{table}[htb]
    \centering
    \caption{Hyperparameter sensitivity analysis for Proactive Drafting's $K$-ary factor ($K$) and tree depth ($D$). All experiments use the full method with $L=64$.}
    \label{tab:ablation_hyperparams}
    \setlength{\tabcolsep}{2.12em}
    \begin{tabular}{c c c}
        \toprule
        \textbf{$K$} & \textbf{$D$} & \textbf{Step Compression} (↑) \\
        \midrule
        2            & 3            & $4.23\times$                  \\
        3            & 3            & $4.42\times$                  \\
        5            & 3            & $4.47\times$                  \\
        \cmidrule(lr){1-3}
        4            & 2            & $4.34\times$                  \\
        4            & 4            & $4.49\times$                  \\
        \cmidrule(lr){1-3}
        \textbf{4}   & \textbf{3}   & $\mathbf{4.51\times}$         \\
        \bottomrule
    \end{tabular}
\end{table}

\begin{figure*}[htb]
    \centering
    \includegraphics{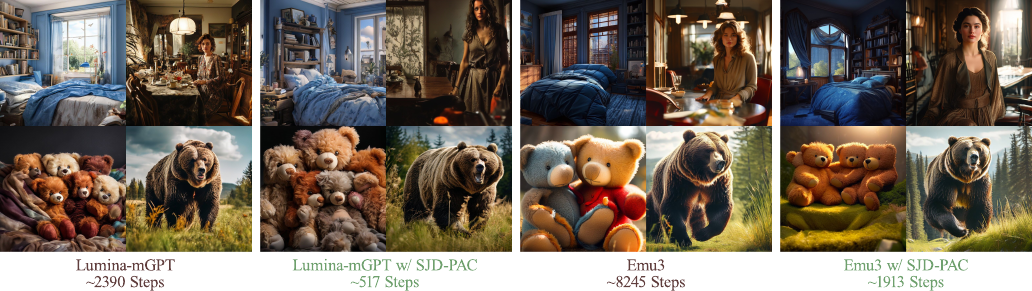}
    \caption{\textbf{Qualitative Comparison.} Images generated using various prompts from the MS-COCO dataset~\cite{coco}. The outputs from our accelerated method (\textcolor[HTML]{6fa16f}{\method}) are visually indistinguishable from those of the baseline models, corroborating our lossless guarantee.}
    \label{fig:qualitative}
\end{figure*}

\subsection{Visual Inspection}
\label{sec:visual_inspect}
Because \method is algorithmically lossless, it theoretically guarantees the preservation of image quality. This is visually supported by samples generated from the MS-COCO dataset~\cite{coco} using Lumina-mGPT~\cite{lumina} and Emu3~\cite{emu3}. As shown in \cref{fig:qualitative}, images produced by our method (labeled \textcolor[HTML]{6fa16f}{w/ \method}) are visually indistinguishable from those generated by the original  models. \method drastically reduces decoding steps while strictly maintaining high-fidelity details.

\begin{figure}[htb]
    \centering
    \includegraphics{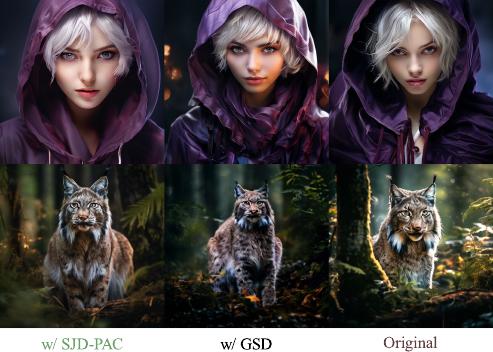}
    \caption{Qualitative comparison with lossy acceleration.}
    \label{fig:vis_case_cmp}
\end{figure}

In contrast, while lossy acceleration methods such as GSD~\cite{gsd} perform well in simple scenarios (\cref{fig:vis_case_cmp}, top row), they introduce noticeable artifacts due to the loss of high-frequency information (bottom row). This highlights the necessity of a lossless approach for preserving fine-grained details during complex image generation.
\section{Conclusion}

In this work, we introduced \method, a novel and strictly lossless framework that directly addresses the inefficiencies of SJD \cite{sjd} through two synergistic mechanisms. First, Proactive Drafting (PD) mitigates frequent single-token acceptance failures by proactively constructing a $K$-ary tree of diverse proposal paths immediately following a rejection, thereby increasing the probability of a valid draft in the subsequent iteration. Second, Adaptive Continuation (AC) refines the SJD verification loop. By continuing verification past the initial mismatch and resampling only the rejected tokens, AC preserves the stability of the sequence tail.

Our experiments demonstrate that \method establishes a new state-of-the-art for training-free, lossless T2I acceleration. The combination of PD and AC reshapes the acceptance distribution, shifting its mass from inefficient short runs to highly efficient long ones. This synergy culminates in a \textbf{4.51$\times$} step compression and a \textbf{3.80$\times$} wall-clock speedup on Lumina-mGPT \cite{lumina}. Notably, \method achieves acceleration competitive with, or even superior to, several prominent lossy methods \cite{lantern++, gsd, sjd2}, proving that substantial speedups in T2I synthesis do not require compromising image fidelity.

While \method is model-agnostic, its performance gains rely on a delicate empirical balance between the Proactive Drafting parameters ($K$, $D$) and the window length ($L$). Future work will explore integrating our framework with orthogonal approaches. Because \method optimizes the drafting and verification processes, pairing it with methods that optimize the acceptance criterion, such as GSD \cite{gsd}, offers a promising direction. A hybrid approach combining our stable, multi-draft verification with advanced acceptance heuristics could push the boundaries of T2I model acceleration even further.
{
  \small
  \bibliographystyle{ieeenat_fullname}
  \bibliography{main}
}


\clearpage
\setcounter{page}{1}
\maketitlesupplementary
\appendix

\section{Proofs of Correctness}
\label{app:proofs}

In this section, we provide the theoretical guarantee that \method is lossless. That is, the distribution of the generated sequence converges exactly to the target distribution $p(x)$, identical to standard autoregressive sampling. We prove this by demonstrating that neither Adaptive Continuation (AC) nor Proactive Drafting (PD) introduces bias into the verified sequence.

\subsection{Preliminaries}
Let $p(x_i \mid x_{<i})$ denote the target distribution at position $i$ given prefix $x_{<i}$, and $q(x_i \mid x_{<i})$ denote the draft distribution. Standard speculative decoding relies on the property that for a random variable $\mathsf{X}$ sampled via the rejection sampling scheme:
\begin{equation}
    \begin{aligned}
        P(\mathsf{X} = x) & = q(x) \cdot \min\left(1, \frac{p(x)}{q(x)}\right)                            \\
                          & \quad + \mathcal{N} \cdot \frac{\max\left(0, p(x) - q(x)\right)}{\mathcal{N}} \\
                          & = p(x),
    \end{aligned}
    \label{eq:rejection_sampling}
\end{equation}
where $\mathcal{N} = \sum_z \max\left(0, p(z) - q(z)\right)$ is the normalization factor for the residual distribution. We refer to this standard result as the \textbf{Rejection Sampling Lemma}.

\subsection{Proof of Correctness for Adaptive Continuation (AC)}

\paragraph{Theorem 1.} \emph{AC preserves the target distribution $p(x)$ for all tokens accepted in the subsequent iteration.}

\begin{proof}
    Let $X^t$ denote the draft sequence generated by the AC mechanism during iteration $t$. At iteration $t+1$, we verify the draft tokens $x_j^t$ against the target distribution $p(x) = p(x \mid X_{<j}^{t+1})$. Let $q(x)$ denote the distribution used to generate the draft token at iteration $t$. Although $q$ may be derived from stale context, it is fixed and known during the verification at $t+1$.

    To prove that the marginal probability of outputting a token matches the target distribution $p(x)$, we analyze the rejection sampling mechanism step-by-step. The probability of outputting a specific token $x$ is the sum of the probability of it being accepted from the draft and the probability of it being resampled after a rejection.

    Let $P(\mathsf{X}^{t+1}_j = x)$ be the total probability that token $x$ is generated at position $j$. We express this as:
    \begin{equation}
        \label{eq:rej_samp}
        P(\mathsf{X}^{t+1}_j = x) = P(\text{Acc. } x) + P(\text{Rej.}) \cdot P(\text{Resamp. } x)
    \end{equation}

    \paragraph{Probability of Acceptance:}
    The draft proposes $x$ with probability $q(x)$. The acceptance probability is defined as $\alpha(x) = \min\left(1, \frac{p(x)}{q(x)}\right)$. Thus, the joint probability of proposing and accepting $x$ is:
    \begin{equation}
        \begin{aligned}
            P(\text{Acc. } x) & = q(x) \cdot \min\left(1, \frac{p(x)}{q(x)}\right) \\
                              & = \min\Big(q(x), p(x)\Big).
        \end{aligned}
    \end{equation}

    \paragraph{Probability of Rejection.}
    The probability of rejecting the draft (summing over all possible vocabulary tokens $v$) corresponds to the probability mass where the draft exceeds the target (or conversely, the missing mass):
    \begin{equation}
        \begin{aligned}
            \label{eq:rej}
            P(\text{Rej.}) & = 1 - \sum_{v \in V} P(\text{Acc. } v)                          \\
                           & = \sum_{v \in V} p(v) - \sum_{v \in V} \min\Big(q(v), p(v)\Big) \\
                           & = \sum_{v \in V} \left( p(v) - \min\Big(q(v), p(v)\Big) \right) \\
                           & = \sum_{v \in V} \max\Big(0, p(v) - q(v)\Big).
        \end{aligned}
    \end{equation}

    \paragraph{Probability of Resampling.}
    Upon rejection, we sample from the residual distribution $p_{\text{res}}(x)$. This distribution is defined as the normalized difference between the target and the draft:
    \begin{equation}
        \begin{aligned}
            P(\text{Resamp. } x) & = p_{\text{res}}(x)                                                                 \\
                                 & = \frac{\max\Big(0, p(x) - q(x)\Big)}{\sum_{v \in V} \max\Big(0, p(v) - q(v)\Big)}.
        \end{aligned}
    \end{equation}
    Observe that the denominator is the normalization constant, which equals $P(\text{Rej.})$ derived in \cref{eq:rej}.

    \paragraph{Combining terms.}
    Substituting these back into \cref{eq:rej_samp} for total probability:
    \begin{equation}
        \begin{aligned}
            P(\mathsf{X}^{t+1}_j = x)
             & = \min\Big(q(x), p(x)\Big)                                                 \\
             & \quad + P(\text{Rej.}) \frac{\max\Big(0, p(x) - q(x)\Big)}{P(\text{Rej.})} \\
             & = \min\Big(q(x), p(x)\Big)                                                 \\
             & \quad + \max\Big(0, p(x) - q(x)\Big)
        \end{aligned}
    \end{equation}

    \noindent We consider two cases to solve this summation for any token $x$:
    \begin{itemize}
        \item \textbf{Case 1}, $p(x) \ge q(x)$: \\
              $\min(q(x), p(x)) + \max(0, p(x) - q(x)) = q(x) + (p(x) - q(x)) = p(x)$.
        \item \textbf{Case 2}, $p(x) < q(x)$: \\
              $\min(q(x), p(x)) + \max(0, p(x) - q(x)) = p(x) + 0 = p(x)$.
    \end{itemize}

    In both cases, the marginal probability $P(\mathsf{X}^{t+1}_j = x)$ is exactly $p(x)$. Crucially, this equality holds independently of the quality or parameters of the draft distribution $q(x)$, provided that $q(x)$ is a valid probability distribution over the vocabulary.
\end{proof}

\subsection{Proof of Correctness for Proactive Drafting (PD)}
\label{app:md_proof}

\paragraph{Theorem 2.} \emph{PD, which involves the sequential verification of $K$ candidates sampled without replacement and is supplemented by sampling from the residual distribution upon total rejection, guarantees exact recovery of the target distribution $p(x)$.}

\begin{proof}
    We prove this theorem by induction on the number of candidates $K$.

    \paragraph{Definitions.}
    Let $p(x)$ denote the target probability distribution and $q(x)$ denote the draft distribution. PD generates a set of candidates by sampling iteratively from $q$ without replacement.
    For the $k$-th candidate $c_k$, let $q^{(k)}(x)$ be the draft distribution adjusted for previous samples $\mathcal{C}_{<k} = \{c_1, \dots, c_{k-1}\}$, and $p^{(k)}(x)$ be the residual target distribution after previous rejections (with $p^{(1)} = p$).
    The adjusted draft distribution is:
    \begin{equation}
        q^{(k)}(x) = \begin{cases}
            \frac{q(x)}{1 - \sum_{y \in \mathcal{C}_{<k}} q(y)} & \text{if } x \notin \mathcal{C}_{<k} \\
            0                                                   & \text{if } x \in \mathcal{C}_{<k}
        \end{cases}
    \end{equation}
    The residual transition follows the Rejection Sampling Lemma established in \cref{eq:rejection_sampling}:
    \begin{equation}
        p^{(k+1)}(x) = \frac{\max\big(0, \; p^{(k)}(x) - q^{(k)}(x)\big)}{\sum_v \max\big(0, \; p^{(k)}(v) - q^{(k)}(v)\big)}.
    \end{equation}

    \paragraph{Base Case ($K=0$).}
    No candidates are proposed. The algorithm samples directly from the residual $p^{(1)}$, which is initialized as $p^{(1)} = p$. The output distribution is trivially correct.

    \paragraph{Inductive Step.}
    Assume that at step $k$, the target distribution is correctly captured by the residual $p^{(k)}$. We verify that the procedure at step $k$ preserves this distribution. Specifically, the probability of rejecting candidate $c_k$ is the complement of the acceptance mass:
    \begin{equation}
        \label{eq:md_rej_prob}
        \begin{aligned}
            P(\text{Rej. } c_k) & = \sum_x q^{(k)}(x) \left(1 - \min\left(1, \frac{p^{(k)}(x)}{q^{(k)}(x)}\right)\right) \\
                                & = \sum_x \max\Big(0, \; p^{(k)}(x) - q^{(k)}(x)\Big).
        \end{aligned}
    \end{equation}

    We now show that the marginal probability of accepting a token $u$ at step $k$, or rejecting it and subsequently sampling it from the next residual $p^{(k+1)}$, equals the probability of sampling $u$ from the current target $p^{(k)}$.
    \begin{equation}
        \begin{aligned}
            p^{(k)}(u) & \stackrel{?}{=} \underbrace{q^{(k)}(u) \cdot \min\left(1, \frac{p^{(k)}(u)}{q^{(k)}(u)}\right)}_{\text{Accept } u \text{ at step } k} \\
                       & \quad + \underbrace{P(\text{Rej. } c_k) \cdot p^{(k+1)}(u)}_{\text{Sample } u \text{ from residual}}
        \end{aligned}
    \end{equation}

    Substituting the definition of $p^{(k+1)}(u)$ and the rejection probability from \cref{eq:md_rej_prob}:
    \begin{equation}
        \begin{aligned}
            \text{RHS} & = \min\left(q^{(k)}(u), \; p^{(k)}(u)\right)                                             \\
                       & \quad + p^{(k+1)}(u) \sum_x \max\left(0, \; p^{(k)}(x) - q^{(k)}(x)\right)               \\
                       & = \min\left(q^{(k)}(u), \; p^{(k)}(u)\right)                                             \\
                       & \quad + \frac{\max\left(0, \; p^{(k)}(u) - q^{(k)}(u)\right)}{\sum_x \dots} \sum_x \dots \\
                       & = \min\Big(q^{(k)}(u), \; p^{(k)}(u)\Big)                                                \\
                       & \quad + \max\Big(0, \; p^{(k)}(u) - q^{(k)}(u)\Big)                                      \\
                       & = p^{(k)}(u).
        \end{aligned}
    \end{equation}

    By induction, since the procedure preserves the target distribution at each step $k$, and the final sampling upon total rejection is drawn from $p^{(K+1)}$, the overall Proactive Drafting process exactly recovers $p(x)$.
\end{proof}

\section{Overhead Analysis}
\label{app:overhead_analysis}

In this section, we analyze the memory consumption and computational latency introduced by \method compared to the vanilla AR baseline.

\paragraph{GPU Memory Consumption.}
We evaluate the memory footprint on Lumina-mGPT \cite{lumina}. The baseline implementation requires approximately $18.7$\,GB of GPU memory. Implementing \method with a decoding window length of $L=64$ results in a memory usage of $18.7$\,GB, demonstrating negligible memory overhead.

Although \method employs a significantly larger decoding window ($L=64$) compared to vanilla AR ($L=1$), the memory cost remains stable. This is because the KV cache is inherently required to store the history of the generated sequence. In our approach, the accepted tokens within the window are immediately converted into persistent KV cache entries. Consequently, the memory used by the window effectively acts as pre-allocated space for the upcoming KV cache, rather than an add-on burden.

\paragraph{Computational Latency.}
We break down the runtime costs per step to analyze the temporal overhead. The breakdown is summarized in \cref{tab:time_overhead}.
For the vanilla AR model, a single forward pass takes approximately $40.8$\,ms, with miscellaneous overheads totaling roughly $1.0$\,ms.
With \method ($L=64$), the model forward pass increases slightly to $44.3$\,ms. The additional components specific to our algorithm include tree verification and generation ($3.2$\,ms) and KV cache maintenance ($2.2$\,ms).

\begin{table}[h]
    \centering
    \caption{Runtime breakdown of a single decoding step in milliseconds. Comparison between Vanilla AR and \method with window length $L=64$.}
    \label{tab:time_overhead}
    \setlength{\tabcolsep}{.17em}
    \begin{tabular}{lcc}
        \toprule
        \textbf{Operation}              & \textbf{Vanilla AR} & \textbf{\method} \\
        \midrule
        Model Forward                   & $40.8$\,ms          & $44.3$\,ms       \\
        Tree Verification \& Generation & -                   & $3.2$\,ms        \\
        KV Cache Update \& Backtracking & -                   & $2.2$\,ms        \\
        Other Overheads                 & $1.0$\,ms           & $1.0$\,ms        \\
        \bottomrule
    \end{tabular}
\end{table}

Despite the overhead of \method, verification is efficient due to the high parallelizability enabled by the static tree structure and independent positional operations. Moreover, as current KV cache overheads arise mainly from non-contiguous memory access, future optimizations for contiguous layouts are expected to further mitigate latency.

\end{document}